\documentclass[runningheads]{llncs}
\usepackage{graphicx}
\usepackage{tikz}
\usepackage{comment}
\usepackage{amsmath,amssymb} % define this before the line numbering.
\usepackage{color}
\usepackage{multirow}
\usepackage{bm}

\usepackage[accsupp]{axessibility}  % Improves PDF readability for those with disabilities.
\usepackage{soul}
\usepackage{booktabs}
\usepackage{makecell}

\begin{document}
% \linenumbers
\pagestyle{headings}
\mainmatter
\def\ECCVSubNumber{6477}  % Insert your submission number here

\title{Incremental Few-Shot Learning via Implanting and Compressing} % Replace with your title

%\begin{comment}
\titlerunning{ECCV-22 submission ID \ECCVSubNumber} 
\authorrunning{ECCV-22 submission ID \ECCVSubNumber} 

\begin{comment}
\author{
	Yiting Li, %\IEEEmembership{Member,~IEEE,}
	Haiyue Zhu, %\IEEEmembership{Student Member,~IEEE,}
	Xijia Feng,
	Jun Ma,
	Cheng Xiang, 
	Prahlad Vadakkepat,
	and Tong Heng Lee\\
	\thanks{Y. Li, X. Feng, C. Xiang, P. Vadakkepat, and T. H. Lee are with the Department of Electrical and Computer Engineering, National University of Singapore, Singapore 117583 (e-mail: yiting\_li@u.nus.edu, fengxijia98@u.nus.edu,  elexc@nus.edu.sg, prahlad@nus.edu.sg, eleleeth@nus.edu.sg).}
	\thanks{H. Zhu is with the
Singapore Institute of Manufacturing Technology, Agency for Science, Technology and Research, Singapore 138634 (zhu\_haiyue@simtech.a-star.edu.sg).}
\thanks{J. Ma is with the Robotics and Autonomous Systems Thrust, The Hong Kong University of Science and Technology, Guangzhou, China, and the Department of Electronic and Computer Engineering, The Hong Kong University of Science and Technology, Clear Water Bay, Kowloon, Hong Kong SAR, China (e-mail: jun.ma@ust.hk).}
}
\end{comment}
\institute{Paper ID \ECCVSubNumber}
%\end{comment}
%******************

\titlerunning{Incremental Few-Shot Learning via Implanting and Compressing}

% visualize parameter space displacement
% visualize training error to show the issue of under-fitting
\author{Yiting Li\inst{1} \and
Haiyue Zhu\inst{2}\thanks{Corresponding Author} \and
Xijia Feng\inst{1} \and
Zilong Cheng\inst{1} \and
Jun Ma\inst{3} \and
Cheng Xiang\inst{1} \and
Prahlad Vadakkepat\inst{1} \and
Tong Heng Lee\inst{1}}
\authorrunning{Y. Li et al.}
\institute{National University of Singapore, Singapore 117583\\
\email{\{yiting\_li,fengxijia98,zilongcheng\}@u.nus.edu,  \{elexc,prahlad,eleleeth\}@nus.edu.sg} \and
SIMTech, Agency for Science, Technology and Research, Singapore 138634\\
\email{zhu\_haiyue@simtech.a-star.edu.sg}
\\
\and
Hong Kong University of Science and Technology, Hong Kong SAR, China\\
\email{jun.ma@ust.hk}}

%******************
\maketitle

\begin{abstract}

This work focuses on tackling the challenging but realistic visual task of Incremental Few-Shot Learning (IFSL), which requires a model to continually learn novel classes from only a few examples while not forgetting the base classes on which it was pre-trained. Our study reveals that the challenges of IFSL lie in both inter-class separation and novel-class representation. Due to intra-class variation, a novel class may implicitly leverage the knowledge from multiple base classes to construct its feature representation. Hence, simply reusing the pre-trained embedding space could lead to a scattered feature distribution and result in category confusion. To address such issues, we propose a two-step learning strategy referred to as \textbf{Im}planting and \textbf{Co}mpressing (\textbf{IMCO}), which optimizes both feature space partition and novel class reconstruction in a systematic manner. Specifically, in the \textbf{Implanting} step, we propose to mimic the data distribution of novel classes with the assistance of data-abundant base set, so that a model could learn semantically-rich features that are beneficial for discriminating between the base and other unseen classes. In the \textbf{Compressing} step, we adapt the feature extractor to precisely represent each novel class for enhancing intra-class compactness, together with a regularized parameter updating rule for preventing aggressive model updating. Finally, we demonstrate that IMCO outperforms competing baselines with a significant margin, both in image classification task and more challenging object detection task.

\end{abstract}

\section{Introduction}
%Humans exhibit remarkable ability in continual adaptation and learning new tasks throughout their lifetime while maintaining the knowledge gained from past experiences. This ability, often known as incremental learning or lifelong learning, is crucial for many real-world applications such as autonomous driving and robotics, which are required to sequentially learn and deal with multiple tasks when implemented in the dynamically changing environment. In the literature, most of the incremental learning (IL) approaches learn new tasks from large scale training samples with annotations. However, massive amounts of labeled data could be prohibitive or expensive to obtain, especially when the new classes are rare categories which are costly or difficult to collect. In practice, it is more realistic that only very small amounts of samples with annotations are available. Therefore, in this paper, we focus on incremental few-shot learning, a more difficult paradigm that aims to continually learn new tasks with only a few examples.
Humans exhibit remarkable ability in continually learning new knowledge while maintaining the knowledge gained from past experiences. This ability, often known as incremental learning or lifelong learning, is crucial for many real-world applications in dynamically changing environments. In the literature, most of the incremental learning approaches~\cite{EWC,IMM,lwf} learn new tasks from large-scale training samples with full supervision. However, massive amounts of labeled data could be prohibitive or expensive to obtain, especially for the real-world applications that often encounter the low-data regime. In practice, it is more realistic that only very small amounts of samples with annotations are available. Therefore, we focus on Incremental Few-Shot Learning (IFSL), a more difficult paradigm that aims to continually learn new tasks with only a few examples.

%For networks are typically trained with non-stationary data streams, one major challenge is catastrophic forgetting, which generally refers to an abrupt performance decrease on previously learned tasks as new tasks are learned. The problem is rooted in the general optimization methods that are being used to encode input data distribution into the parametric representation of the network during training. Due to the inaccessibility to previous data while learning a new task, gradient-based optimization methods will change the learned encoding to minimize the objective function with respect to the current data distribution. For example, in complex incremental learning scenarios new tasks might not be tightly linked to previous tasks, the training process on one task will discard information that was irrelevant for the task at hand, but that would be relevant for the previous tasks, hence leading to forgetting.

For networks trained with non-stationary data streams, one major challenge is the catastrophic forgetting, which generally refers to an abrupt performance degradation on previously-learned tasks after adapting to new tasks. Such problem is rooted into the gradient-based optimization methods that are commonly used for training deep networks. Since a network will encode the input distribution into its learnable parameters through gradient descent, when the previous examples are inaccessible or the new tasks are not tightly linked to previous ones, minimizing the objective function to fit the current data distribution will discard the information that was irrelevant for the current task, but such information could be important for previous tasks. Moreover, since gradients induced from small training set could be highly biased with low diversity, the optimization trajectory of training loss will easily fall into sharp and narrow local minima~\cite{keskar2016large}, which irreversibly results in overfitting.

In the literature, existing works mainly focus on learning unbiased classifiers through either meta \cite{Once,zhang2021hallucination} or transfer learning strategies \cite{DFS,earth,topic,gandi}. Those approaches heavily rely on the pre-trained knowledge gained from the large-scale base set, i.e., the pre-trained embedding space is fully reused by freezing backbone parameters, and only the linear classifier is learned during the incremental learning. The advantage of these approaches is that the few-shot fine-tuning can be treated as a convex optimization problem, which are less sensitive to overfitting and catastrophic forgetting. However, we argue that these methods still have adverse effects in view of feature space separation and novel class representation. Due to intra-class variation,  the network will automatically associate a novel class with one or a few of base classes to construct its feature space \cite{fad,li2021beyond}. As a result, the obtained feature manifold of a novel class will have an incompact intra-class structure that scatters across feature clusters of its nearby base classes, hence leading to classification confusion.% \hl{(Can use a TSNE plot to demo this?)}

In IFSL the objective is to discriminate all classes (base and novel) from all learned tasks. However, since the model cannot access to old classes when adapting to a new one, the resultant network weights cannot optimally discriminate all classes. To address this dilemma, we propose a novel IFSL framework, termed as \textbf{Im}planting and \textbf{Co}mpressing (\textbf{IMCO}). For the first pre-training stage, different from the closed-set pre-training strategies that implicitly learn transferable features by discriminating a limited number of base classes, we propose to additionally learn semantically-rich features that could be beneficial for discriminating between the base and other unseen classes. To achieve this goal, we propose an open-set~\cite{open1,open2,open3} pre-training method that explicitly mimics the data distribution of novel classes with the assistance of data-abundant base classes. By doing so, the model will have early access to the upcoming novel class during pre-training and encode their knowledge into parameters, as shown in Figure.\ref{fig:intro}(a). 

Motivated by the fact that features extracted by a pre-trained model for data from a novel class are usually entangled with its nearby base classes \cite{nearest}, we propose to artificially synthesize novel classes that lie between the feature space of any two base classes.  The synthesized novel classes could be interpreted as anchors in embedding space that efficiently anticipates the distribution of real-world unseen classes. Therefore, we refer the proposed pre-training strategy as \textbf{Implanting}, since it allocates feature space for constructing unseen class before they become available. The synthesized novel-class instances could push the feature space of any two base classes to be far away from each other, which facilitates better separation between base and novel classes, as shown in Figure.\ref{fig:intro}(b).
\begin{figure*}[tb!]
	\centering
	\centering{\includegraphics[width=1.0\columnwidth]{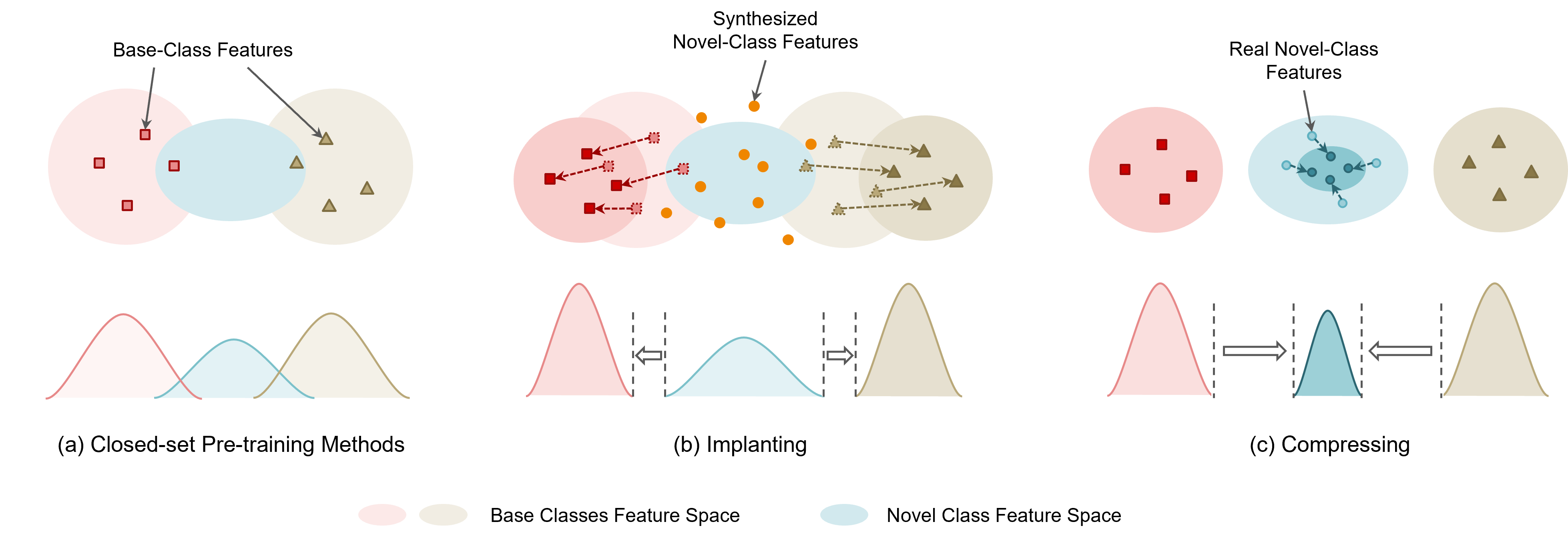}}
	\caption{(a) Under the closed-set pre-training paradigm, the model learns decision boundaries that segments the feature space into several non-overlapping regions occupied by each base class. Due to the intra-class variation, the pre-trained model will automatically associate a novel class with its similar base classes, thus leading to a scattered feature representation. (b) In the implanting step, we explicitly mimic the distribution of novel classes by mixing features of any two base classes. The synthesized instances could push the feature space of any two base classes to be far away from each other, which facilitates feature space partition. (c) In the compressing step, we further fine-tune backbone for precisely representing novel classes in feature space. By optimizing intra-class compactness, the newly learned classification boundaries are less sensitive to the choice of few-shot samples. }
	\label{fig:intro}
\end{figure*}

However, merely optimizing inter-class separability is still not sufficient for obtaining good few-shot performance. The features of an unseen class must be clustered tightly to prevent the learned decision boundaries from being instance-dependent. Therefore, our second intuition is to cluster each novel class more tightly in feature space when their data become available, so that the new classification boundaries learned during incremental learning are less sensitive to the choice of few-shot samples. To achieve this goal, we further fine-tune higher feature extraction layers for precisely representing them in feature space. This step is referred as \textbf{Compressing}, as we fine-tune the feature extractor to encourage the embedding vectors to move quickly towards their corresponding classification weight vectors (class centroids), as shown in Figure.\ref{fig:intro}(c). 

%To address such an issue, we propose a new learning framework for robust and efficient fine-tuning during sequential model adaption. The proposed framework contains two important ingredients:

%, hence forming tighter clusters of each novel classes
%to ensure the intra-class compactness
%In particular, when features of novel class become spread out, the classification boundaries formed by sampling one-shot data often misclassify large regions. In contrast, as features in a class are compacted, the dependence of the class boundary on the choice of one-shot samples becomes weaker.

%Rather than attempting to constrain the distance between the all pre-trained and fine-tuned weights in the network using a single projection, constraints are applied on a layer-wise basis. This makes optimisation more manageable and also allows practitioners to favour fine-tuning certain parts of the network

Since the data of old tasks is inaccessible and the new-task samples are few-shot, naively fine-tuning without effective regularization will lead to aggressive updating \cite{jiang2019smart} and result in both overfitting and forgetting. Regarding to this, our study suggests that we should update the model only within a small neighborhood of the previous task's  weights to stabilize the fine-tuning process \cite{gouk2020distance,aghajanyan2020better}. To accomplish this goal, we introduce a projected-subgradient-type regularization at each iteration, which constrains the adaption process to search within a set of weights within a predefined distance from the starting point. Moreover, we also constrain the updating of important network parameters so that the categorical information of old tasks could be well preserved. Experiments confirm that our approach is particularly beneficial for preventing aggressive updating, because it guarantees that the enforced constraints could be fulfilled even if one uses a large number of iterations to train the network parameters. 

%While learning new tasks, the network weights relevant to old tasks are updated to minimize a loss on the new taskà As a result, performance on previous tasks suffers often dramaticallyà

%Weight regularization. The rst class of approaches focuses on preventing weight drift determined to be relevant for previous tasksà They do so by estimating a prior importance of each parameter in the network which are assumed to be independent after learning each taskà When training on new tasks, the importance of each parameter is used to penalize changes to themà

To this end, the main contribution of this study can be summarised as follows: (1) By explicitly mimicking the feature distribution of novel classes during pre-training, the proposed implanting method provides a model with a better initialisation such that learning can modify the weights less in the later incremental stages. (2) The proposed compressing method facilitates smooth feature-level adaption by enforcing constraints on the weight-space displacement, hence enhancing intra-class compactness and further enlarging the inter-class separability between all classes.

\section{Related Works}

\textbf{Incremental Learning}
A common line of research on preventing catastrophic forgetting is the regularization-based incremental learning (IL) methods, when adapting to new task, they constrain the model’s output logits or important parameters corresponding to the previous tasks by using an extra regularization term in loss function. For example, EWC~\cite{EWC} penalizes the changes to the important parameters. IMM~\cite{IMM} performs a separate merging step after learning each new task. PackNet~\cite{mallya2018packnet} employs heuristic weight pruning through a binary mask to isolate important model parameters. LwF~\cite{lwf} use the previous model output as soft labels to constrain the model being trained on the new data. While they could be somehow helpful, IFSL is far more than a straightforward application of these methods due to the additional challenge of overfitting. 
%From our observation, they often suffer from performance degradation when learning with few supervisions, i.e., good performance can only be maintained for the first few tasks, then rapidly decays in later tasks. 
%This is because, they only constrain new task's parameter updating direction to not interfere with previous tasks, i.e., gradients produced on the current task should align with the feasible directions outlined by previous task, but not strictly restricting the magnitude of overall parameter displacement to be sufficient small. As a result,  the overwhelming number of less important parameters could be updated dramatically, and the target model will be inevitably driven far away from the starting point, which may fail to intersect with previous tasks’ local minima which are already sharp and narrow. 

\textbf{Few-Shot Learning}
Few-shot learning aims at reocognizing novel visual concepts through only a few training samples. However, most of the existing approaches merely focus on the simple n-way k-shot classification setting, instead of discriminating all learned classes. Modern few-shot learning can be divided into two categories: gradient-based approaches~\cite{Chelsea_2017_MAML,Lee_2018_MLAL} and metric-based approaches~\cite{Xue_2020_OneShot,Snell_2017_Prototypical,Sung_2018_RLFW}. Gradient-based approaches aim at effectively adapting a model to fit unseen categories through a limited number of parameter updates. For example, Model-Agnostic Meta-Learning (MAML)~\cite{Chelsea_2017_MAML} encodes transformable meta knowledge into network initialization, where a small number of gradient updates can lead to good performance. Metric learning~\cite{Snell_2017_Prototypical} learns a nearest-neighbor embedding function, which encodes the same-category objects into a compact cluster in the feature space. 

\textbf{Incremental Few-Shot Learning}
In the literature, it has recently been investigated that the classifier learning strategies, which reuse the pre-trained embedding space and merely synthesis classification weights for adding novel classes~\cite{imtfa,Once,earth,topic,gandi,zhang2021hallucination}. By freezing the backbone parameters, they are less prone to issue of catastrophic forgetting caused by continually evolved training sets. However, as the base classes have no overlap with the novel ones, features required to classify novel classes might be quite different from the base classes. As a result, completely freezing the backbone network without any form of fine-tuning could result in the issue of intransigence \cite{chaudhry2018riemannian}, where a model tends to be overly stable and lacks the plasticity necessary to integrate new knowledge.

\section{Our Approach}
In this section, we first review the preliminaries of IFSL setting and the conventional two-stage training pipeline, then we introduce our method that tackles IFSL via implanting and compressing.
\subsection{Preliminaries}
The IFSL problem is generally formulated as a two-phase learning task. In the pre-training phase ($t=0$), a model is first trained on a large-scale and category-abundant base set for a generalizable feature representation. In the incremental learning phase ($t>0$), continual model adaption is performed over multiple stages to adapt into a sequence of tasks, where each task contains $m$ novel classes with only a few training samples. When adapting to a new class, we assume that training samples of old classes are not accessible for joint fine-tuning. Finally, the model is evaluated on test samples from all encountered classes.

%Classifier learning methods are the current one of the leading paradigms for IFSL. Aiming at preventing overfitting and forgetting during fine-tuning, only the fully-connected classification layer, are updated to fit the few-shot samples. While the feature extractor is frozen to preserve the pre-trained knowledge on the abundant base classes.

%Although the current development of classifier learning strategies could bring considerable achievements, their performance heavily rely on the similarity between the base and novel classes, i.e., The higher the similarity, the higher the efficiency. Therefore, when there are no strong similarities between the base and novel domains, the pre-trained embedding space could be imperfect for representing novel classes and thus result in a scattered feature representation. 

\textbf{Problem of the closed-set pre-training strategy} One major challenge of the first-stage pre-training is that: How to make the feature extractor learn image features that are more generalizable to data-rare novel classes?  Most of existing works naively assume a closed-world setting with a limited number of base classes, while seldom caring about what properties are owned by the learned representations and whether they are indeed beneficial for separating base and novel instances. In particular, we assume that there are $m$ distinct classes contained in a data-abundant base set. Under the closed-set training setting~\cite{openmax,openworldobject}, the base model is enforced to learn a separable feature space with decision boundaries that segments the feature space into $m$ non-overlapping regions. However, due to the absence of novel classes during pre-training on base set, a model could only learn features that are relevant to differentiate base classes, which might not be descriptive enough to separate base classes from upcoming novel classes. As a result, for a new class that is disjoint from base set, the pre-trained embedding space will automatically associate it with one or a few of the $m$ base classes according to its intra-class variation, thus leading to a scattered feature representation that are entangled with its similar base classes.

\subsection{Implanting Step}

Towards addressing this drawback, we propose an open-set pre-training strategy named as Implanting, which prepares a closed-set model for extending to more classes. In particular, we artificially augment the base-set training tasks to mimic the emergence of unseen classes. Our intuition here is to force the model to learn a more diversified set of image features that are beneficial for separating base from other unseen classes, and this, in turn, improves its capability to adapt to novel classes with few training data.

In practice, many recent few-shot works assume that the instances of novel classes will share some common properties with base classes. For example, if a model is trained on a base set of birds, a novel set will probably be some other categories of birds from a similar domain. This is a common assumption of few-shot learning which enables effective cross-category generalization. Thus, we adopt the same assumption that novel classes belong to a subspace of a certain domain where base classes exists. Therefore, we could synthesize novel classes by sampling instances from data-abundant base classes, and transferring closed-set training into open-set training.

%which utilizes the semantic knowledge gained from data-abundant base classes to facilitate the learning of data-hungry novel classes.

%By doing so, we could transfer the acquired knowledge from data-rich base set to semantically-similar and yet data-hungry novel set.
%For instance, interpolating between images of two birds will not result in a realistic photo of another bird.
A natural way to achieve this is linear interpolation between raw images. However, since image space is highly nonlinear, such practice could hardly produce reasonable results that align well with real-world image distribution. On another hand, with a well-trained feature extractor, all classes will be encoded by corresponding class prototypes in feature space and each of them will be linear separable from others. Motivated by this, we propose to generate novel-class instances lying between the feature manifold of either two base classes. Therefore, the obtained between-class features could share some common properties with base classes~\cite{verma2019manifold}, thus aligning well with the real-data manifold. 

In particular, we generate a synthesized novel-class sample $\bm{x}_{sn}$ by convexly combining random pairs of feature vectors belonging to any two base classes. In particular, let $\bm{x}_{b+}$ and $\bm{x}_{b-}$ be the embedding vectors from two different base classes, a random ratio $\lambda$ is sampled from Beta distribution $\beta(\cdot,\cdot)$, and the two base feature vectors are mixed to produce the features of a synthesized novel class, by
\begin{equation}
\begin{aligned}
\label{eq:mixup}
\bm{x}_{sn} =  Mix_{  \lambda  } \left( \bm{x}_{b+},\bm{x}_{b-}\right), 
\end{aligned}
\end{equation}
\normalsize
where ${Mix_{ \lambda  } \left(\bm{a},\bm{b}\right)} =  \lambda \cdot \bm{a} + \left( 1 - \lambda \right)\cdot \bm{b}$. In the first stage of base training, we simulate a real incremental process by organizing the base data into episodes for constructing small learning tasks. Consider a training episode $\{\mathcal{S}_{b},\mathcal{Q}_{b}\}$ which consists of $n$ base classes, where $\mathcal{S}_{(\cdot)}$ and $\mathcal{Q}_{(\cdot)}$ denote the support and query sets, respectively. We first artificially generate another $n$ synthesized novel-class training samples $\{\mathcal{S}_{sn},\mathcal{Q}_{sn}\}$ through linear interpolation in Eq.\eqref{eq:mixup}. After that, the support set and query set of the original base classes and synthesized novel classes are concatenated together to form a augmented training episode, i.e., $\mathcal{S}_{aug} = \{\mathcal{S}_{b},\mathcal{S}_{sn}\}$ and $\mathcal{Q}_{aug} = \{\mathcal{Q}_{b},\mathcal{Q}_{sn}\}$. The augmented support set $\mathcal{S}_{aug}$ is then used to learn the classifier weights of different classes, and the query set $\mathcal{Q}_{aug}$ is used to compute training loss for optimizing the feature extractor. Leveraging the interpolation of deep features, the mixed instances could further optimize the decision boundary in the embedding space towards the decision regions of base classes. Hence, the feature representations of base classes would be much tighter, leaving more space for representing the upcoming novel classes.

%Many recent meta-learning algorithms also make use of idea that keeping fine-tuned weights close to their initial values is desirable. However, these approaches typically focus on developing methods for learning the initial weights, rather than working with pre-specified initial weights.
%instead of penalising the distance of the parameters from zero, they are regularised towards a bias vector. This bias vector is learned during the course of solving least squares problems on a collection of related tasks.
%Previous work investigating the generalisation performance of neural network based on the distance the weights have travelled from their initial values has done so with the aim of explaining why existing methods for training models work well.

\begin{figure*}[tb!]
	\centering
	\centering{\includegraphics[width=1.0\columnwidth]{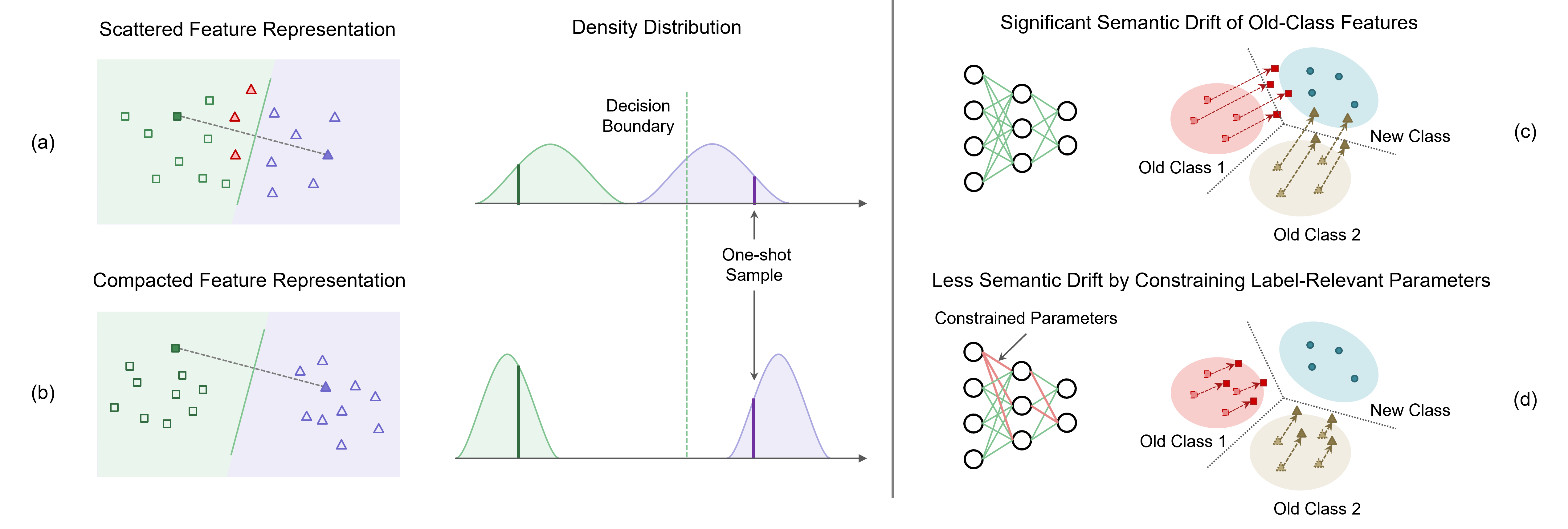}}
	\caption{ (a) When features of a novel class become scattered, the decision boundaries learned from few-shot samples could be biased and with low accuracy. (b) When feature distribution become more compacted, the learned decision boundaries are more robust to the choice of few-shot samples. (c)
	Upon adapting to a new task, modifications to the label-relevant parameters of old classes could cause higher performance loss and result in semantic drift of feature space. (d) To prevent the feature space of old classes from drifting too much, we propose to preserve categorical information by constraining the label-relevant parameters.}
	\label{fig:method}
\end{figure*}
\subsection{Compressing Step}
After the Implanting step,  base classes will have a more compact intra-class distribution and become more separable from the upcoming novel classes. However, merely ensuring linear separability is not sufficient for achieving good few-shot performance. In particular, as shown in Figure.\ref{fig:method}(a), when features of a novel class become scattered, the decision boundaries learned from few-shot samples could be biased and with low accuracy. In contrast, as shown in Figure.\ref{fig:method}(b), when feature distribution become more compacted, the learned decision boundaries are more robust to the choice of few-shot samples. Therefore, clustering each novel class more tightly is highly desired when their examples are rare, so that the new classification boundaries learned are less sensitive to the choice of few-shot samples.

%To prevent aggressive updating, our study suggests that we should update the model only within a small neighborhood of the previous task's  weights to stabilize the fine-tuning process. To accomplish this goal, we introduce a projected-subgradient-type regularization at each iteration, which constrains the adaption process to search within a set of weights within a predefined distance from the starting point. Experiments confirm that our approach is particularly beneficial for preventing aggressive updating, because it guarantees that the constraints will be fulfilled even if one uses a large number of iterations to train the network parameters. 
%\subsubsection{Connection of Loss Landscape and IFSL}
Towards this goal, we further fine-tune higher feature extraction layers upon the arrival of each novel class. However, due to data scarcity, naively fine-tuning without effective regularization will cause aggressive model updating and result in overfitting. To address such issue, we propose a projected-subgradient-type regularization that constrains the optimization trajectory of learning each task to be within a small sphere centered on the previous task model weights. In particular, our intuition is twofold. 

1) \textbf{Constraining Weight-Space Changes}: To maintain the low-error region on old tasks when adapting to a new task, the sequentially learned local minima are desired to overlap with each other in loss landscape. Since networks fine-tuned with rare data tend to overfit and end up with sharp minima, the weight-space displacement between the sequential tasks should be constrained more strictly in order to facilitate intersection. To achieve the above goals, we propose a Dense Model Fusion (DMF) mechanism that could be easily integrated into any gradient-based optimizer. In particular, we perform weight-space model fusion densely after each time of parameter updating, i.e., after every training iteration. We prove that the weight displacement of model parameters could be tightly constrained by using our method, which is particularly beneficial for preventing aggressive updating.

2) \textbf{Maintaining Categorical Information}: When fine-tuning a network to update its weights to fit a new task, modifications to the label-relevant parameters of old classes could cause higher performance loss comparing with the label-irrelevant ones \cite{HAT,rain}, and resulting in semantic drift of previous classes' representation \cite{incrementalreview}, as shown in Figure.\ref{fig:method}(c). For label-relevant parameters, we refer to the parameters that encode categorical information so that the classification tasks can be accurately performed. In contrast, label-irrelevant parameters usually encode more general appearance information such as pose, shape, or other characteristics (depth and illumination, etc.). To prevent the feature space of old classes from drifting too much from their optimal locations when adapting to new classes,  we propose to preserve categorical information by constraining the label-relevant parameters through effective regularization. In other words, we allow the label-irrelevant parameters to be re-used and adapted for precisely representing novel classes, as shown in Figure.\ref{fig:method}(d).

%The forgetting which is caused by the large weight and activation drift originated from the large domain shifts seems to dominateà
%representations of previous data from drifting too much while learning new tasks

%In summary, our key insight is what we call ”unactivated channel re-usage”. Specifically our approach identifies those transferable channels and preserves such filters through regularization and identify those untransferable channels and reuse them, using an attention mechanism with feature map regularization.

%From a loss landscape view, the training loss of old classes are sharp and unstable along the side of label-relevant parameters, but maintaining flat and convex along the side of the label-irrelevant ones as they are more transferable \cite{valley}. Therefore, the label-relevant parameters should be constrained to be less updated when adapting to new classes, so that the optimisation trajectories of a new task could be restricted within the flat side of previous tasks' local minima, hence remaining low error on old classes. Towards this, we develop an efficient method to automatically determine the label-relevance of model parameters for old tasks through adversarial weight-space perturbation. 

\subsubsection{Parameter Importance Evaluation}

%an independent branch for each new task while freezing the old task parameters to preserve the old knowledge
Intuitively, a parameter with greater capacity of categorical information are usually more important for maintaining low training-set error. Therefore, we propose to evaluate the label relevance of model parameters through \textbf{adversarial weight perturbation}. In particular, we can find the worst case by maximizing the classification loss of the network on the old task once an incremental session is finished. Formally, assuming that model trained on the $t$-th task $\mathcal{D}_t$ is denoted as $W_t$, the label relevance of the solution $W_t$ in the old parameter space can be predicted by maximizing the following adversarial loss function,

\begin{equation}
\begin{aligned}
&\max_{W_t} L_{adv}
=&\sum\nolimits_{i=1}^{n} \Big( L_{ce}\big(f(x_i;W_t),y_i\big)+ D_{kl}\big(P(x_i;W_t)||P(x_i;W_{t-1})\big)\Big)
 \end{aligned}
\end{equation}
\normalsize
where $f(x;W)$ denotes a mapping function $f$ associated with the parameter $W$ from input image $x$ to an output space, $L_{ce}$ represents the standard cross entropy loss for classifying images in $\mathcal{D}_t$, which is used to evaluate categorical information of the current training set. For the categorical information of the previous $t-1$ tasks, we utilize knowledge distillation $D_{kl}(P||Q)$ from the previous model $W_{t-1}$ (trained on the $(t-1)$ th task) to the current model $W_{t}$ to approximate, where $D_{kl}(P||Q) = P\cdot log(P/Q)$ denotes the KL-divergence of two distributions $P$ and $Q$. 

Upon training finished, we further fine-tune $W_t$ with the above adversarial loss $L_{adv}(W_t,W_{t-1})$ for an additional epoch to obtain the  perturbed model $W_t^{adv}$. Then we compare $W_t^{adv}$ with the original model $W_t$ to calculate label relevance of each parameter. In particular, let $K$ denote the number of layers in the model, and $l_k$ denotes a certain layer which contains $N$ parameters, where $l_k = \{p_1^k,p_2^k, ...,p_N^k\}$. Here, $p_j^k$ denotes the $j$-th parameter of layer $l_k$, and we use $^{adv}p_j^k$ to denote the value of the same parameter in the adversarially trained model $W_t^{adv}$. Then we compute the adversarial weight perturbation for the $k$ th layer, where ${V_k} = \{v_1^k,v_2^k ...v_N^k\}$ and $v_j^k = | ^{adv}p_j^k - p_j^k|^{2}$.

The parameters that yield significant change of values are more sensitive to the adversarial loss, hence containing more categorical information than the others. Next, we further normalize $v_j^k$ within each layer $l_k$ to form parameter importance $I_k$, where $I_k = \{v_1^k,v_2^k ...,v_N^k\}/{\max (V_k)} $.
Finally, the obtained importance matrix $I_k$ is accumulated along the historical tasks for better remembering all seen tasks. For example, given the accumulated importance matrix for the previous $t-1$ tasks $S_k^{t-1}$ in the $k$ th layer, the formal importance matrix $S_k^{t}$ is computed as
\begin{equation}
\begin{aligned}
S_k^{t} =\min\left((I_k^t +S_k^{t-1}),1\right),
\end{aligned}
\end{equation}
\normalsize
where $\min(\cdot\,,1)$ is to bound the importance value by 1 element-wisely. The recursive update of the importance matrix allows us to aggregate the memory about all past tasks, by involving one new task at a time.

\subsubsection{Dense Model Fusion (DMF)}

In the DMF algorithm, the importance matrix $S^t$ is the key for overcoming catastrophic interference in sequential learning. The detailed procedures for the proposed DMF method upon adapting to the $(t+1)$ th task is detailed as follows:

\begin{enumerate}
  \item Initialization of linear classifier: the weight vector of each class is initialized by averaging feature embeddings in the training set $\mathcal{D}_{t+1}$, the obtained model is denoted as $W_{init}$.
  \item Forward propagate the inputs of the $n$-th iteration, then back propagate the errors and calculate weight modifications $\Delta W_n = \eta \cdot \nabla L_n$ by the standard SGD optimizer, and update the weight matrix in each layer by
\begin{equation}
\label{eq:1}
\begin{aligned}
W_{n} =W_{n-1} -  \Delta W_n.
\end{aligned}
\end{equation}
\normalsize
  \item Performing model fusion with the starting point $W_{init}$, using the importance information $S^t$ as projection matrix, where $\alpha$ is a damping ratio for controlling training plasticity, and $\textbf{1}$ denotes the matrix that has the same size as $S^t\alpha$ with all entries as 1.
\begin{equation}
\begin{aligned}
W_{n}  \leftarrow W_{n}\cdot (\textbf{1}-S^t \alpha )  + W_{init} \cdot S^t \alpha,
\label{eq:fusion}
\end{aligned}
\end{equation}
\normalsize
  \item Repeat steps (2) to (3) for the next batch.
  \item Upon finishing of training, calculate the parameter importance $S^{t+1}$ based on $\mathcal{D}_{t+1}$.
\end{enumerate}

Next, we analyze the proposed DMF method mathematically to show that the magnitude of weight displacement for each task is tightly upper-bounded and controllable. In particular, substituting Eq.~\ref{eq:1} into Eq.~\ref{eq:fusion} recursively, we will have:
\begin{equation}
\label{eq:3}
\begin{aligned}
W_{n} = W_{init} - \left((\textbf{1}-S^t\alpha)^n\cdot \Delta W_1+(\textbf{1}-S^t\alpha)^{n-1}\cdot \Delta W_2+...(\textbf{1}-S^t\alpha)^1\cdot \Delta W_n\right)
\end{aligned}
\end{equation}
\normalsize
and the magnitude of weight displacement can be calculated as
\begin{equation}
\label{eq:4}
\begin{aligned}
|W_{n}-W_{init}| = | \sum\nolimits_{i = 1}^{n} (\textbf{1}-S^t\alpha)^i\cdot \Delta W_{n-i+1}|\leq \sum\nolimits_{i = 1}^{n} |(\textbf{1}-S^t\alpha)|^i\cdot |\Delta W_{n-i+1}|
\end{aligned}
\end{equation}
\normalsize
Assuming that the per-iteration parameter displacement $\Delta W$ is upper-bounded by $\Delta W_{max}$, which could be implemented through setting clipping threshold for the gradient $\nabla L$. As a result, the upper-bound of weight space displacement could be calculated as
\begin{equation}
\label{eq:5}
\begin{aligned}
 \lim_{n\to  \infty } |W_{n}-W_{init}| =  \lim_{n\to  \infty }  |\Delta W_{max}| \sum\nolimits_{i = 1}^{n} |(\textbf{1}-S^t\alpha)|^i
%\\= \lim_{n\to  \infty }  |\Delta W_{max}|\cdot |S_i^T\alpha| \frac{ (1-|S_i^T\alpha|^{n}) }{(1-|S_i^T\alpha|} 
=   |\Delta W_{max}| \frac{ 1 -|S^t\alpha| }{|S^t\alpha|}.
\end{aligned}
\end{equation}
\normalsize
This proves that parameter displacement during sequential adaption is tightly constrained even if one uses a large number of iterations to train the network parameters, which facilitates the intersection of local minima and alleviates catastrophic forgetting. For the damping ratio $\alpha$, when we set $\alpha=0$, the optimization process will degrade to conventional fine-tuning and results in aggressive updating.  In contrast, when we increase $\alpha$ such that $|S_i^t\alpha|=1$, the new model is kept to be same as the old one and suffers from intransigence (under-fitting), since the previous learned feature space might not be representative enough for understanding the current task.

%In case where the new task is highly relevant to the old ones, or aggregated partial knowledge obtained from each task is sufficient to explain the new task, a small searching space could be sufficient for finding the optimal parameters of new task. However, when the learned features cannot accurately represent the new task, we should accordingly enlarge the searching space so that more capacity can be released. To this end, we develop an error-driven approach that could dynamically adjust the network capacity by increasing or decreasing the radius of searching space according to the training error of each iteration, hence achieving a better trade-off between stability and plasticity,
\textbf{Error-Guided Plasticity-Stability Adjustment}
With a pre-defined learning rate, the magnitude of weight displacement will mainly depend on the damping ratio $\alpha$, i.e., a smaller damping ratio allows larger searching space of parameter adaption. To maintain sufficient capacity for new task learning, we argue that the size of searching space should be dynamically adjusted according to task difficulty. In case where the new-coming classes could be well represented by the previous learned embedding space, a small searching space could be sufficient for finding the optimal parameters of new task, hence a larger $\alpha$ is adopted to avoid unnecessary parameter updates. In contrast, when the previous learned features cannot accurately represent the new task, we should elastically expands model capacity by allowing more searching space for new-task adaption. 

To this end, we develop an error-driven approach that could dynamically adjust the network capacity by increasing or decreasing the radius of searching space according to the training error $e$ at each iteration. Precisely, the adaptive damping ratio $\alpha$ for each training iteration is calculated as,
\begin{equation}
\begin{aligned}
\alpha = r-  \mu_{\alpha} \cdot e,
\end{aligned}
\end{equation}
where $r$ is the basic damping ratio and $\mu_{\alpha}$ is a scalar coefficient. By doing so, plasticity is emphasized for hard tasks to achieve better adaption, while stability is emphasized for early tasks to prevent aggressive fine-tuning.

\section{Experiments}

\subsection{Dataset Setting}
We conduct extensive experiments on two popular image classification datasets mini-ImageNet and CUB200, and one object detection dataset MS-COCO to show the effectiveness and universality of the proposed methods.

\textbf{Mini-ImageNet} 
The mini-ImageNet dataset contains 100 classes which contain 500 images for training and 100 images for testing in each class. Following the common practice, the 100 classes are divided into 60 base classes and 40 novel classes. To perform incremental learning, the 40 novel classes are further divided into 8 incremental sessions with 5 classes in each session, where each class are only provided with 5 examples.

\textbf{CUB200}
The CUB200 dataset is commonly used for fine-grained image classification, which contains 200 classes of birds. The images are resized to 256 × 256 and then cropped to 224 × 224 for training. We split the 200 classes into 100 base classes and 100 novel classes, and then dividing 100 novel classes into 10 incremental sessions, where each novel classes is provided with 5 training examples. 

\textbf{MS-COCO}
We split the 80 categories into two groups, 20 classes which are overlapped with Pascal VOC are used as novel classes, and the remaining 60 classes are used as base classes. The 20 novel classes are then divided into 10 or 20 incremental sessions with one class in each session. During incremental model adaption, the model can only access $k = \{5, 10\}$ bounding-box annotations of the current-task classes. Finally, we evaluate our model on all the 80 classes.

\subsection{Implementation Details}
Our implementation is built with PyTorch library, and the code for reproducing experiment results can be found in supplemental materials.

%\hl{(detail number of $\beta \left( 1.5,  1.5 \right)$ should put in Implementation Details)}
\textbf{Image Classification}
We employ ResNet18 for experiments on mini-ImageNet and CUB200. At the implanting stage, we first train our network
through SGD with learning rate of 0.0002 and momentum of 0.9, and we use $\beta \left( 1.5,  1.5 \right)$ for mixing two base classes. In the compressing stage, we select the last two convolution layers to fine-tune, because top layers usually encode semantic-rich feature representations. Random crop, random scale, and random horizontal flip are employed for augmentation during training. For the adaptive damping ratio $\alpha = r-  \mu_{\alpha} \cdot e$ used in our error-guided plasticity-stability adjustment method, we set $r=0.3$ and $\mu_{\alpha}=0.4$.

\textbf{Object Detection}
We use Faster-RCNN as detection architecture and Resnet-50~\cite{ResNet} as backbone. The detection model is pre-trained on base set for the first 8 epochs, with a batch size of 16, learning rate of 0.01, momentum of 0.9, and weight decay of 0.0001. Next, we fine-tune the pre-trained model with a sequence of disjoint sets of novel classes. A smaller learning rate of 0.005 is adopted for stabilizing training since there are fewer examples. Different from image classification, the training error $e$ for object detection is calculated based on the foreground RoIs ($IOU>0.5$).

\begin{table*}[h]
\renewcommand\arraystretch{1.25}
\scriptsize
%\caption{\textbf{Effectiveness of Each Component}}

\caption{\textbf{Effectiveness of Each Component.} Averaged over 5 runs. \textbf{Metric}: Classification Accuracy, \textbf{the higher the better}.}
\centering
% \resizebox{\linewidth}{!}{
\begin{tabular}{m{0.9cm}<{\centering} m{0.9cm}<{\centering} m{0.9cm}<{\centering} cccc ccccccccc}
% \begin{tabular}{cccc ccccccccc}
\toprule
\multirow{2}*{\textbf{IM}} & \multirow{2}*{\textbf{MF}} & \multirow{2}*{\textbf{IPP}} & \multirow{2}*{\textbf{EGPSA}} & \multicolumn{9}{c}{\textbf{Acc. in each session(\%)}}  \\
\cline{5-13}
~ & ~ & ~ & ~ &  \textbf{0} &  \textbf{1}  & \textbf{2} &  \textbf{3} & \textbf{4} &  \textbf{5} & \textbf{6} &  \textbf{7} & \textbf{8}

\\
%& \textbf{72.00} & \textbf{66.83} & \textbf{62.97} & 59.43 & 56.70 & 53.73 &51.19&49.24 &47.63 
\midrule
& & &            & 70.36  & 64.76  & 59.66  & 55.59  & 53.18  & 48.92 & 45.67  & 43.28  & 41.39
\\
\checkmark & &  &  & 71.35  & 66.54  & 62.72  & 59.21  & 56.68  & 53.63 & 51.07  & 49.23  & 47.55
\\
\checkmark &\checkmark &  &  & 71.35  & 66.51  & 62.86  & 59.38  & 56.74  & 53.70 &51.09  & 49.35  & 47.72
\\
\checkmark &\checkmark &\checkmark & & 71.35 & 66.73 & 62.91 & 59.53 & 56.84 & 53.90 & 51.26  & 49.52  & 47.89
\\
\checkmark &\checkmark &\checkmark &\checkmark & \textbf{71.35} & \textbf{66.78} & \textbf{62.95} & \textbf{59.56} & \textbf{56.91} & \textbf{53.86} & \textbf{51.28} & \textbf{49.53} & \textbf{47.95} 
\\
\bottomrule
\end{tabular}
% }
\label{table:ablation}
\vspace{-4mm}
\end{table*}

\subsection{Analysis}
In this part, we design various experiments to evaluate the effectiveness of our algorithm and study the characteristics of different components.The experiments are performed on mini-ImageNet with ResNet18.

\textbf{Ablation study.} We conduct the following ablation studies to investigate the contribution of each component to the final performance gain. (1) Implanting (\textbf{IM}) : Only the proposed Implanting method is employed to facilitate pre-training. Since the Compressing step is not performed, the feature extractor is frozen when adapting to novel classes, while the classifier weight vector of each new class is initialized by averaging their feature-space embeddings. (2) Model Fusion (\textbf{MF}): Vanilla model fusion is performed in each training iteration, which is a simplified form of Eq.\ref{eq:fusion} without the Parameter Importance matrix $S$.
(3) Important Parameter Preservation (\textbf{IPP}): parameter importance information $S$ is added into Eq.\ref{eq:fusion} for elastically constraining parameter updating. However, we still use a constant value 0.3 for the damping ratio $\alpha$. (4) Error-Guided Plasticity-Stability Adjustment (\textbf{EGPSA}): We fine-tune the damping ratio according to the training error, which dynamically releasing more capacity for handling hard tasks. Regarding the results presented in Table~\ref{table:ablation}, we have several observations. 

1) Simply adopting the closed-set pre-training setting leads to the poor generalization performance on novel classes due to its ineffectiveness in addressing category confusion (first line of results). Training with the augmented novel classes could effectively alleviate such issue, outperforming the naive pre-training approach by 6.56 points. 

2) Fine-tuning feature extractor with vanilla model fusion brings additional performance improvement than freezing it. Moreover, utilizing parameter importance information could achieve up to 0.34 points performance gain than using the vanilla model fusion, which verifies the effectiveness of the proposed adversarial weight perturbation in identifying label-relevant parameters. It also shows that solving the challenging IFSL problem requires both constraining weight-space displacement and maintaining important categorical information.

3) EGPSA further improves the classification performance in all incremental sessions. It demonstrate the effectiveness of dynamically adjusting parameter searching space for improving the training flexibility.

\begin{figure*}[tb!]
	\centering
	\centering{\includegraphics[width=0.8\columnwidth]{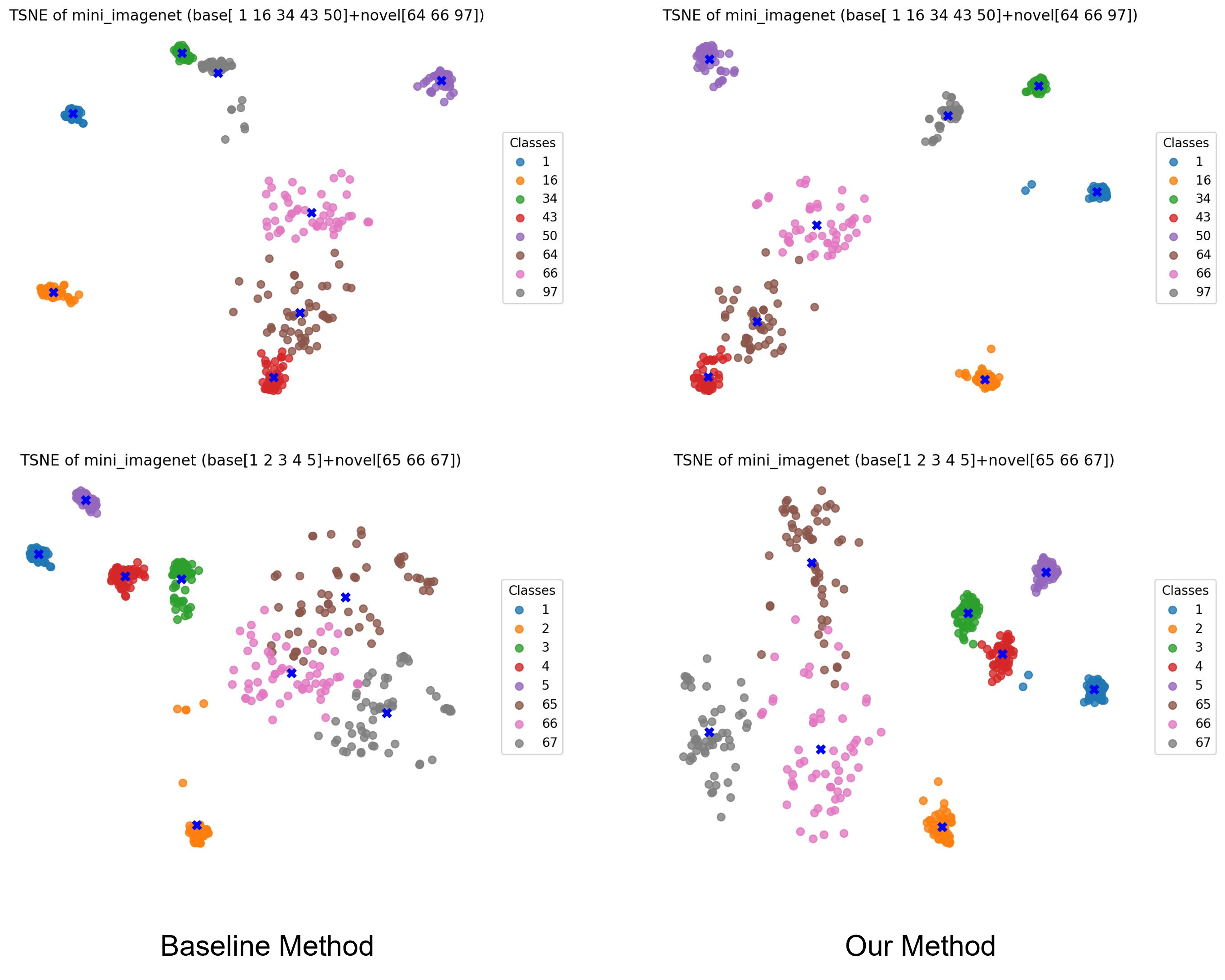}}
	\caption{(a)Baseline method: Embedding space learned by the closed-set pre-training strategy.  (b) Our method: Embedding-space learned by open-set \textbf{implanting} method. }
	\label{fig:intro}
\end{figure*}

\begin{figure*}[tb!]
	\centering
	\centering{\includegraphics[width=0.9\columnwidth]{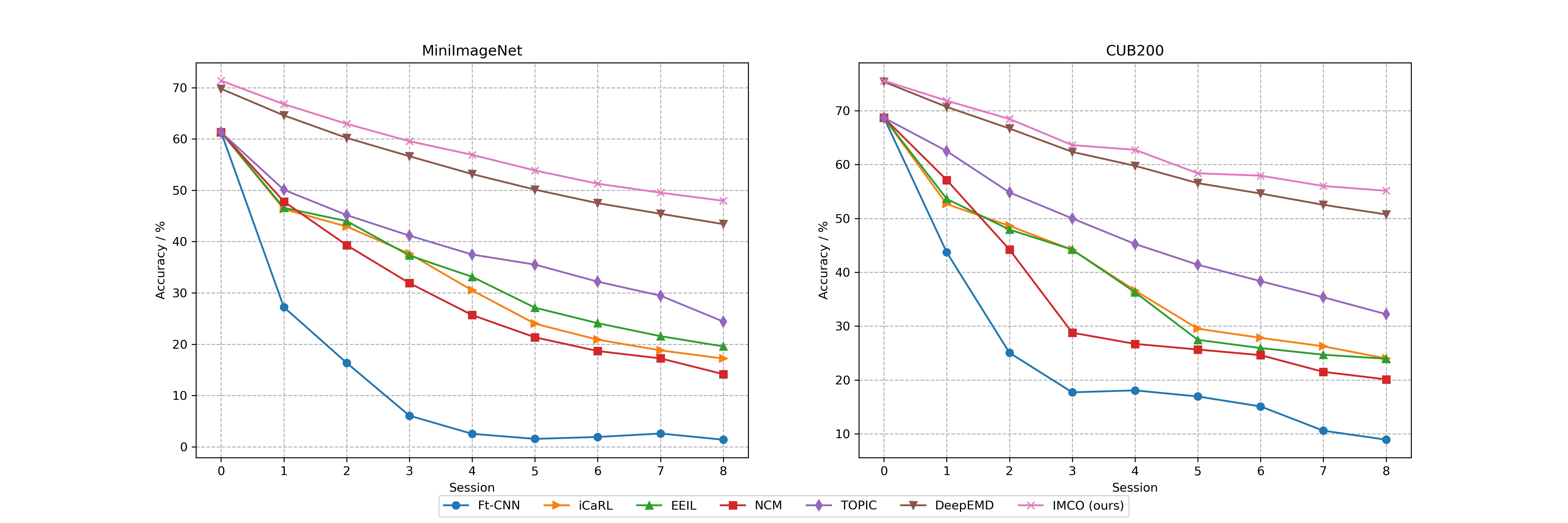}}
	\caption{ (a) 5-shot 8-incremental steps on Mini-Imagenet. (b) 5-shot 10-incremental steps on CUB200.}
	\label{fig:result}
\end{figure*}

\begin{table*}[h]
\scriptsize
% \tiny
\renewcommand\arraystretch{1.25}
\caption{\textbf{Performance on Mini-ImageNet} averaged over 5 runs. \textbf{Metric}: classification accuracy, \textbf{the higher the better}.}
\centering
% \resizebox{\linewidth}{!}{
\begin{tabular}{l c c c cccccccccccc}

\toprule
\multirow{2}*{\textbf{Method}} & \multicolumn{9}{c}{\textbf{Acc. in each session(\%)}} 
\\
\cline{2-10}
& \textbf{0}& \textbf{1}& \textbf{2}  & \textbf{3}& \textbf{4} & \textbf{5}& \textbf{6} & \textbf{7}& \textbf{8}  

\\
\midrule
Ft-CNN & 61.31& 27.22  & 16.37    & 6.08     & 2.54  & 1.56 &1.93 &2.6 &1.4    
\\
iCaRL~\cite{rebuffi2017icarl} & 61.31& 46.32 & 42.94    & 37.63     & 30.49 & 24.00 &20.89 &18.8 &17.21        
\\
EEIL~\cite{eeil} & 61.31& 46.58 & 44.00     & 37.29     & 33.14 & 27.12 &24.1 &21.57 &19.58      
\\
NCM~\cite{ncm} & 61.31& 47.8 & 39.31     & 31.91     & 25.68 & 21.35 &18.67 &17.24 &14.17        
\\
TOPIC~\cite{topic}& 61.31 & 50.09 & 45.17     & 41.16    & 37.48 & 35.52 & 32.19 &29.46 &24.42     
\\
%LFFS~\cite{vinyals2016matching}& .37 & 65.45 & 61.41     & 58.00     & 54.81 & 51.89 &49.10 &47.27 &45.63    
%\\
DeepEMD~\cite{earth} & 69.77 & 64.59 & 60.21     & 56.63    & 53.16 & 50.13 &47.49 &45.42 &43.41    
\\
%CEC~\cite{continually} & \textbf{72.00} & \textbf{66.83} & \textbf{62.97} & 59.43 & 56.70 & 53.73 &51.19&49.24 &47.63 
%\\
\midrule
\textbf{IMCO} & \textbf{71.35} & \textbf{66.78} & \textbf{62.95} & \textbf{59.56} & \textbf{56.91} & \textbf{53.86} & \textbf{51.28} & \textbf{49.53} & \textbf{47.95} 
%\\
%\textbf{IMCO+CEC} & \textbf{72.08} & \textbf{67.01} & \textbf{63.12} & \textbf{59.87} & \textbf{56.95} & \textbf{54.06} & \textbf{51.64} & \textbf{49.77} & \textbf{48.04} 
\\
\bottomrule
\end{tabular}
% }
\label{table:imagenet}
\vspace{-2mm}
\end{table*}

\begin{table*}[h]
\scriptsize
% \tiny
\renewcommand\arraystretch{1.25}
\caption{\textbf{Performance on CUB200}. }
\centering
% \resizebox{\linewidth}{!}{
\begin{tabular}{l c c c cccccccccccc cc}

\toprule
\multirow{2}*{\textbf{Method}} & \multicolumn{9}{c}{\textbf{Acc. in each session(\%)}} 
\\
\cline{2-12}
& \textbf{0}& \textbf{1}& \textbf{2}  & \textbf{3}& \textbf{4} & \textbf{5}& \textbf{6} & \textbf{7}& \textbf{8}  & \textbf{9}& \textbf{10} 

\\
\midrule
Ft-CNN & 68.68& 43.7  & 25.05   & 17.72    & 18.08  & 16.95 &15.1 &10.6 &8.93    & 8.93 &8.47
\\
iCaRL~\cite{rebuffi2017icarl} & 68.68& 52.65 & 48.61   & 44.16    & 36.62 & 29.52 &27.83 &26.26 &24.01&23.89&21.16       
\\
EEIL~\cite{eeil} & 68.68& 53.63 & 47.91    & 44.2     & 36.3 & 27.46 &25.93 &24.7 &23.95&24.13&22.11      
\\
NCM~\cite{ncm} & 68.68& 57.12 & 44.21     & 28.78    & 26.71 & 25.66 &24.62 &21.52 &20.12&20.06&19.87       
\\
TOPIC~\cite{topic}& 68.68 & 62.49 & 54.81    & 49.99    & 45.25 & 41.4 & 38.35 &35.36 &32.22&28.31&26.28    
\\
%LFFS~\cite{vinyals2016matching}& .37 & 65.45 & 61.41     & 58.00     & 54.81 & 51.89 &49.10 &47.27 &45.63    
%\\
DeepEMD~\cite{earth} & 75.35 & 70.69 & 66.68    & 62.34    & 59.76 & 56.54 &54.61 &52.52 &50.73&49.20&47.60   
\\
%CEC~\cite{continually} & \textbf{72.00} & \textbf{66.83} & \textbf{62.97} & 59.43 & 56.70 & 53.73 &51.19&49.24 &47.63 
%\\
\midrule
\textbf{IMCO} & \textbf{77.56} & \textbf{73.14} & \textbf{69.10} & \textbf{64.53} & \textbf{63.21} & \textbf{59.86} & \textbf{58.43} & \textbf{56.72} & \textbf{55.65} & \textbf{54.05} & \textbf{52.45} 
%\\
%\textbf{IMCO+CEC} & \textbf{72.08} & \textbf{67.01} & \textbf{63.12} & \textbf{59.87} & \textbf{56.95} & \textbf{54.06} & \textbf{51.64} & \textbf{49.77} & \textbf{48.04} 
\\
\bottomrule
\end{tabular}
% }
\label{table:imagenet}
\vspace{-2mm}
\end{table*}

%\textbf{Visualization of feature space.} We visualize the image embeddings and classifier weights by t-SNE. We randomly choose five classes from Mini-ImageNet as the base classes, and we add another five novel classes that are unseen from pre-training. Comparing with the baseline, the proposed IMCO method significantly improves feature-space compactness and separability, hence generating better decision boundaries that are more robust to the randomness of few-shot examples.

\subsection{Comparison with SOTA Methods}
We then compare our performance with the state-of-the-art results on two classification benchmarks:  Mini-Imagenet and CUB200. We show the results in Fig.~\ref{fig:result} and the detailed results on each incremental session in Table. \ref{table:imagenet}. Comparing with the existing classifier learning methods, the proposed IMCO method significantly improves feature-space compactness and separability, hence generating better decision boundaries that are more robust to the randomness of few-shot examples. To this end, we achieve best results over all sessions with a significant margin of 4.5 points. Moreover, since our methods focus on representing learning without resulting in any extra computational burden, which is also friendly for applications deployed on low-power hardware such as edge devices.

\textbf{Visualization of feature space.} We visualize the image features by t-SNE. We randomly choose five classes from Mini-ImageNet as the base classes, and we add another three novel classes that are unseen from pre-training. Comparing with the baseline, the proposed implanting method significantly improves feature-space compactness and separability, i.e., classes are more separable with our implanting method, hence generating better decision boundaries that are more robust to the randomness of few-shot examples. This indicates the necessity of employing an open-set learning setting during pre-training.
\begin{table*}[h]
\scriptsize
% \tiny
\caption{Results on MS-COCO in comparison to state-of-the-art methods, averaged over 5 runs. \textbf{Metric}: Performance represents in Multi-Class Average Precision (mAP). \textbf{Novel}: mAP on novel classes \textbf{All}: mAP on all classes. \textbf{The higher the better}.}

\centering
% \vspace{1mm}
% \resizebox{\linewidth}{!}{
\begin{tabular}{l | cccc | cccc }
\toprule
\multirow{2}*{\textbf{Method}} &    \multicolumn{4}{c|}{\textbf{20-task}} & \multicolumn{4}{c}{\textbf{10-task}}  \\
&\multicolumn{2}{c}{\textbf{5-shot}} & \multicolumn{2}{c|}{\textbf{10-shot}} 
&\multicolumn{2}{c}{\textbf{5-shot}} & \multicolumn{2}{c}{\textbf{10-shot}} 

\\
\midrule
&Novel & All & Novel & All 
&Novel & All
&Novel & All
 \\
\midrule

FRCN-ft
&       1.2 &  4.60
&       1.6 &  5.90

&       2.3 & 10.7
&       3.5 & 10.4

\\
Meta-RCNN~\cite{Yan2019MetaRCNN}
&       2.8 &  10.7
&       3.3 &  11.1

&       3.1 &  11.6
&       4.3 &  13.4

\\
MetaDet~\cite{wang2019meta}
&       4.2  & 18.1 
&       5.3  & 18.3 

&       4.7  & 19.2 
&       5.5  & 19.6

\\
%CEC~\cite{continually}
%&       4.9  & 18.6 
%&       5.5  & 18.7 

%&       5.3  & 19.3 
%&       6.0  & 19.3

%\\
TFA~\cite{Wang2020}
&       3.4  & 18.4 
&       4.9  & 18.5 

&       4.3  & 18.4 
&       5.1  & 18.5 

\\
IMTFA~\cite{imtfa}
&       4.2  & 18.5 
&       5.5  & 18.6

&       4.3  & 18.3 
&       5.5  & 18.6 

\\

ILOD~\cite{incrementalOBJECTDETETION}
&       3.6  & 11.3      
&       4.3  & 11.1

&       4.1 & 12.5
&       5.5 & 13.8

\\

EWC~\cite{EWC}
&       3.1 & 10.5
&       4.0 & 10.9    

&       4.7 & 11.8
&       5.3 & 12.1

\\
IMM~\cite{IMM}
&       5.1  & 18.3   
&       5.6  & 19.0

&       4.4 & 18.6
&       6.7 & 19.2

\\

\midrule

\textbf{IMCO}
&       \textbf{7.6} & \textbf{19.0}  
&       \textbf{8.3} & \textbf{19.3}     

&       \textbf{8.1} & \textbf{19.7}
&       \textbf{8.9} & \textbf{20.1}

\\
\bottomrule
\end{tabular}
% }
\label{table:COCO}
\vspace{-3mm}
\end{table*}

\subsection{Extending to Object Detection}
We then compare our performance with the state-of-the-art methods in Few-Shot Object Detection. The first two baselines, Meta-RCNN~\cite{Yan2019MetaRCNN} and MetaDet~\cite{wang2019meta}  belong to meta-learning approaches that predict unseen classes conditioned on a set of support examples. The next two baselines TFA~\cite{Wang2020} and IMTFA~\cite{imtfa} belong to classifier learning methods that reuse the pre-trained backbone for representing novel classes. We also compare with conventional incremental methods (IMM, EWC and ILOD) that are originally developed for data-abundant scenarios, we use the naive fine-tuning method as the baseline, denoting as FRCN-ft, where we unfreeze the ROI feature extractor to fine-tune with novel classes. Then we implement different regularization-based incremental learning methods into the baseline method FRCN-ft.

We test our method in a long sequence of incremental object detection tasks, where the 20 novel classes are divided into 10 or 20 incremental steps. The sequential fine-tuning is conducted by adding one group of classes at each time until all 20 classes are learned. As shown in Table~\ref{table:COCO}, our method outperforms the classifier learning methods (TFA and IMTFA) with a significant margin. This verifies that fine-tuning deep features is more important for real-world detection tasks, which is consistent with the findings in~\cite{sun2021fsce}. Comparing with the incremental learning methods such as EWC and ILOD, enforcing a constraint on the weights by the proposed compressing method is a better regularization strategy than adding a term to the objective function that penalizes the deviations of weights.

%\subsection{Limitations}
%We argue IMCO is not a perfect method for cross-domain few-shot learning. Since the success of IMCO requires sufficient domain similarity between base and novel classes, so that we could still obtain high accuracy on novel classes even when the weight-space change is constrained to be very small. However, when base and novel classes are from different domains, there will be significantly less intersection between their testing local minima, thus merely exploring the neighbor space of the pre-trained weights might not be a feasible method for adapting to novel classes.

%\section{Appendix}

%\clearpage\mbox{}Page \thepage\ of the manuscript.
%\clearpage\mbox{}Page \thepage\ of the manuscript.

%This is the last page of the manuscript.
%\par\vfill\par
%Now we have reached the maximum size of the ECCV 2022 submission (excluding references).
%References should start immediately after the main text, but can continue on p.15 if needed.
\section{Conclusions}
In this work, we propose a generic learning scheme \textbf{IMCO} for addressing the challenging IFSL problem. By explicitly mimicking the feature distribution of novel classes during pre-training, the proposed implanting method provides a model with a generalizable representation that facilitate better separation between base and novel classes. With an effective weight-space regularization, the proposed compressing method facilitates smooth feature-level adaption by preventing semantic drift on label-relevant parameters, which not only enhances intra-class compactness but also prevents aggressive model updating. Moreover, we propose to dynamically adjust model capacity according to task difficulty, which achieves a better trade-off between plasticity and stability.

%\clearpage
\bibliographystyle{splncs04}
\bibliography{eccv2022submission}
\end{document}